# Algorithmic Simplicity and Relevance

Jean-Louis Dessalles

Telecom ParisTech – 46 rue Barrault – F-75013 Paris – France
dessalles@Telecom-ParisTech.fr - www.dessalles.fr

**Abstract.** The human mind is known to be sensitive to complexity. For instance, the visual system reconstructs hidden parts of objects following a principle of maximum simplicity. We suggest here that higher cognitive processes, such as the selection of relevant situations, are sensitive to *variations* of complexity. Situations are *relevant* to human beings when they appear *simpler to describe than to generate*. This definition offers a predictive (*i.e.* falsifiable) model for the selection of situations worth reporting (interestingness) and for what individuals consider an appropriate move in conversation.

## 1  Complexity, Simplicity and the Human Mind

Almost half a century ago, Ray Solomonoff suggested that inductive learning is guided by simplicity [1]. In 1999, Nick Chater drew attention to the fact that several other human cognitive processes are guided by a principle of minimum complexity [2-3]. Figure 1 illustrates the fact that our visual system reconstructs hidden parts of shapes by preferring simplest patterns.

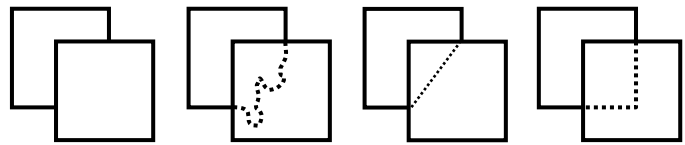

**Fig. 1.** Hidden shapes are as the least complex ones (after [2]).

Surprisingly, the same principle seems to be at work also in higher cognitive processes. Simplicity explains the sensitivity to coincidences [4-5]. It accounts for cognitive biases such as representativeness [6]. It makes predictions about how news elicit emotion, depending on parameters such as proximity, rarity or prominence [7]. Simplicity, defined as the discrepancy between universal probability distribution and uniform distribution, has been used to measure subjective improbability [8]. Complexity has also been related to the feeling of beauty and its variations has been claimed to define interestingness [9]. Complexity is also claimed to be involved in several human traits such as supernatural beliefs, creativity, humor and fiction ([9]; [10] p. 545; [11] p. 967).

Our own work on interestingness led us to develop *simplicity theory* (ST) [7], [12]. The central claim of ST is that subjective probability depends on the difference between generation complexity and description complexity. In these previous developments of ST, interestingness (which is a form of relevance) was merely equated with unexpectedness. The present paper is an attempt to bring the notion of *relevance* closer to an operational definition, based on algorithmic theory, by defining feature relevance and by distinguishing first-order from second-order relevance. In what follows, I will first mention some previous attempts to define relevance, and remind that it is a crucial issue. Then I will briefly outline the central notions of Simplicity Theory, and show how ST can provide a formal definition of relevance. I will then illustrate through examples the explanatory power of the definition. In the conclusion, I will consider the current limits of the theory by mentioning recent work on the impact of emotion on relevance.

## 2      Relevance

Relevance is an empirical phenomenon. Human conversation is a risky game in which speakers dare not misjudge what is worth telling, lest they be punished by being considered socially inept [13]. Being relevant is an essential part of what constitutes human intelligence, as opposed to what we share with other animals. Historically, the main contribution to a theory of relevance was offered by Dan Sperber and Deirdre Wilson [14]. These authors introduce two new notions: cognitive effect and cognitive cost. They define a linguistic utterance as relevant if it maximizes the former and minimizes the latter.

The main merit of this definition is to place the problem on the cognitive ground. Relevance is no longer a question of social convention [15], nor a mere statistical observation about what is generally said and not said in specific contexts. Relevant utterances are supposed to result from genuine computations. Unfortunately, Sperber and Wilson do not provide details about how these computations are performed.

In previous studies, we distinguished two forms of relevance in language, depending on the conversational mode [7]. *Narrative relevance* corresponds to interestingness. A good model of narrative relevance should explain how speakers select events worth reporting in conversational narratives. *Argumentative relevance*, on the other hand, controls the appropriateness of conversational moves during a discussion. Together, conversational narratives and discussions represent more than 90% of conversational time [7].

Algorithmic simplicity may offer a predictive model of narrative relevance [7]. The present paper proposes a few formal definitions about what makes an event relevant. It will also explore how far the same model can be extended to argumentative relevance.

# 3   Simplicity Theory

Simplicity Theory (ST) defines simplicity, not in absolute terms, but as the difference in complexity between expectations and observation. To do so, it distinguishes the standard notion of *description* complexity from *generation* complexity.

The *description complexity* $C(s)$ of a situation $s$ is the length of the shortest description of $s$ that the observer may achieve. This notion coincides with Kolmogorov complexity, but the choice of the machine is not free. The description machine is bound to be the observer, with her previous knowledge and computing abilities.

*Generation complexity* $C_w(s)$ measures expectations about $s$, in complexity terms. It is defined as the length of the minimal program that must be given to the "world" for it to generate $s$. Again, this corresponds to the standard definition of Kolmogorov complexity, except that the machine is bound to function according to what the observer knows about the world's constraints (note that the "world" has no objective character here). In particular, $s$ is supposed to be generated according to some causal process. This constraint affects $C_w(s)$, which may therefore depart from $C(s)$.

The restriction to particular machines, together with a limited time constraint as in [9], makes $C(s)$ and $C_w(s)$ computable. Both computations, however, differ significantly. The description complexity $C(s)$ of a situation that the observer already encountered may be given by its address in memory. If the observer's memory is organized as a binary tree, then $C(s)$ depends on the location of $s$ on that tree. In the computation of $C(s)$ and $C_w(s)$, each bit counts. It is therefore crucial that addresses in memory be minimal in length. A minimal (maximally compact) code is easy to design for lists, using a positional code. Table 1 offers an example of positional code for a list. Note that the code is not self-delimited: this is the price to pay for having a compact code; as a consequence, one must allow the use of a punctuation symbol, or equivalently suppose that code segmentation is performed at a preprocessing stage.

When the observer's memory structure is unavailable, it is sometimes possible to assess the relative complexity of items (objects, people, places…) by comparing their relative ranks on the Web using the number of hits given by a search engine. It is for instance easy to compute the complexity of US presidents, as shown in table 1 (note that G.H. Bush is downgraded, as many pages about him do not mention the middle initial). The code used to compute complexity is designed to be maximally compact (more compact that a prefix-free code). With this code, once the list of US-president is available, its first item, B. Obama, is meant by default and its complexity is zero.

Generation complexity $C_w(s)$ is computed in a different way. The most basic generation machine is a uniform lottery among $N$ objects, in which case $C_w(s) = \log(N)$. When several independent lotteries are used, the complexities add up to give $C_w(s)$. In contrast to the observation machine, the generation machine can be considered to be memory-less. If the same number $n$ comes out twice in a row in a Lottery game, the generation complexity of the double event is $2 \times C_w(n)$, whereas its description complexity is only $C(n)$.

**Table 1.** Web-complexity of the 10 last US presidents, using a non self-delimited code.

| President | Number of hits | Code | Complexity |
|---|---|---|---|
| Barak Obama | 263000000 |  | 0 |
| George W. Bush | 63300000 | 0 | 1 |
| John Kennedy | 57500000 | 1 | 1 |
| Bill Clinton | 46200000 | 00 | 2 |
| Ronald Reagan | 32200000 | 01 | 2 |
| Jimmy Carter | 18200000 | 10 | 2 |
| Richard Nixon | 14200000 | 11 | 2 |
| Lyndon Johnson | 13200000 | 000 | 3 |
| Gerald Ford | 9900000 | 001 | 3 |
| George H. Bush | 6260000 | 010 | 3 |

The central notion of ST is *unexpectedness* $U(s)$. It is defined as:

$$U(s) = C_w(s) - C(s) \qquad (1)$$

In a Lottery game, all draws are supposed to have the same generation complexity $C_w(s) \approx 6 \times \log(49)$ (supposing that 6 numbers are drawn between 1 and 49). When a remarkable combination such as 1–2–3–4–5–6 comes out, description complexity $C(s)$ is much smaller, and $U(s)$ reaches significant value.

Figure 2 illustrates the difference between generation and description. The black ball has to go trough five binary choices before reaching a leaf $s$ of the tree. Therefore, $C_w(s) = 5$. If $s$ is indistinguishable from other leaves, then $C(s) = \log(32) = 5$ and $U(s) = 0$. If $s$ happens to be the only white leaf, then in this case $C(s) = 0$ (as 'being white' is the only apparent feature among leaves) and $U(s) = 5$.

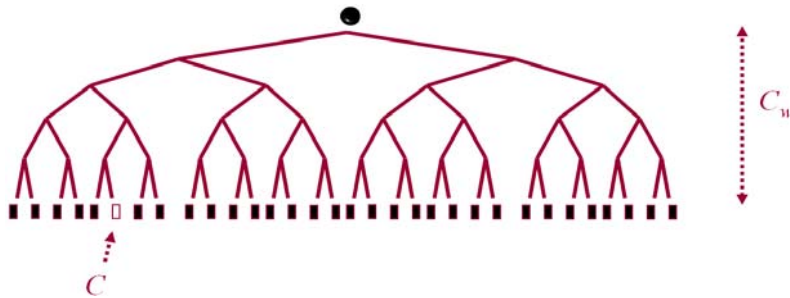

**Fig. 2.** Generation complexity *vs.* description complexity.

More generally, generation complexity is given by the complexity of the simplest causal scenario or theory that may have produced the actual situation (about causality and complexity, see [11], section 7.4). Dowe proposes a computation in which theories are ranked by their complexity ([10] p. 545, note 206): a special list of theories, called 'miracles', is located somewhere in the theory hierarchy; for some observers, for whom miracles are not stored too deep in the complexity-based hierarchy, invok-ing miracles might be a parsimonious way to account for the generation of a given state of affairs. This illustrates the fact that $C_w(s)$ is observer-dependent, as well as description complexity $C(s)$. Whenever these computations are available, relevance can be quantitatively defined.

## 4  Relevance from an algorithmic perspective

We must distinguish two cases. A situation, or a property of a situation, may be relevant because it contributes to making a topic interesting. We call this quality *first-order relevance*. A situation or a property may also be relevant because it makes a previous (relevant) topic *less* relevant. We call this quality *second-order relevance*, or 2-relevance. We define these two notions in turn.

### 4.1  First-order Relevance

Relevance cannot be equated with failure to anticipate [16]: white noise is 'boring', although it impossible to predict and is thus always 'surprising', even for an optimal learner. Our definition of unexpectedness, given by (1), correctly declares white noise uninteresting, as its value *s* at a given time is hard to describe but also equally hard to generate (since a white noise amounts to a uniform lottery), and therefore $U(s) = 0$.

Following definition (1), some situations can be 'more than expected'. For instance, if *s* is about the death last week of a 40-year old woman who lived in a far place hardly known to the observer, then $U(s)$ is likely to be negative, as the minimal description of the woman will exceed in length the minimal parameter settings that the world requires to generate her death. If death is compared with a uniform lottery, then $C_w(s)$ is the number of bits required to 'choose' the week of her death: $C_w(s) \approx \log_2(52 \times 40) = 11$ bits. If we must discriminate the woman among all currently living humans, we need $C(s) = \log_2(7 \times 10^9) = 33$ bits, and $U(s) = 11 - 33 = -22$ is negative. *Relevant* situations are unexpected situations.

$$s \text{ is relevant if } \quad U(s) = C_w(s) - C(s) > 0 \qquad (2)$$

Relevant situations are thus simpler to describe than to generate. In our previous example, this would happen if the dying woman lives in the vicinity, or is an acquaintance, or is a celebrity. Relevance is detected either because the world generates a situation that turns out to be simple for the observer, or because the situation that is observed was thought by the observer to be 'impossible' (*i.e.* hard to generate).

In other contexts, some authors have noticed the relation between *interestingness* and unexpectedness [9, 16], or suggested that the *originality* of an idea could be measured by the complexity of its description using previous knowledge ([10], p. 545). All these definitions compare the complexity of the actual situation *s* to some reference, which represents the observer's expectations. For instance, the notion of *randomness deficiency* ([8], ch. 4 p. 280) compares actual situation to the output of a uniform lottery. The present proposal differs by making the notion of expectation (here: generation) explicit, and by contrasting its complexity $C_w(s)$ with description complexity $C(s)$.

Situations correspond to states of the world. As such, they cannot be grasped in every detail. This is not a problem, however, as we can focus on specific aspects of a given situation. Relevant aspects constitute the essential part of narratives. Consider a feature $f$ that is present in situation $s$. For instance, the fact that a given individual, Ryan, is eating a hot-dog. Considering $f$ as a logical predicate, this means that $f(s)$ is regarded as true. We may write, on the generation side:

$$C_w(s) \leq C_w(f(s)) + C_w(s \mid f(s)) \qquad (3)$$

If Ryan could choose freely among 16 possible sandwiches, the first term $C_w(f(s))$ amounts to $C_w(f(s)) = \log(16) = 4$ bits. However, if we know that Ryan is Muslim, $C_w(f(s))$ can reach significant values as, by default, non-eating pork is a low *mutable* property (see section 5.2) for Muslims. On the description side, we have:

$$C(s) \leq C(f) + C(s \mid f(s)) \qquad (4)$$

In contrast with (3), $f(s)$ needs only to be described through a description of $f$ and not to be generated as a fact. The term $C(f)$ measures the conceptual complexity of $f$ for the observer in the current context. In the example of Ryan's meal, the conceptual complexity of 'hot-dog' can be estimated by the logarithm of the rank of this type of food in a list of typical meals or in short-term memory.

Features are relevant with respect to a given situation if they contribute to unexpectedness.

$$f \text{ is relevant w.r.t. } s \text{ if } \quad U(f(s)) = C_w(f(s)) - C(f) > 0 \qquad (5)$$

Definitions (2) and (5) control what is worth telling when reporting or signaling an event in conversation. Note that if $f$ is the conjunction of several sub-properties, those sub-properties need not be relevant separately. The art of telling narratives is to assemble elements that, together, produce unexpectedness. A conjecture is that every descriptive element, in spontaneous narratives, is intended to make relevance eventually maximal.

## 4.2 Second-order relevance

An admissible reaction to relevant topics consists in attempting to diminish their unexpectedness. The following definition concerns a piece of information $t$ that may alter unexpectedness.

$$\text{if } U(s \mid t) < U(s), \text{ then } t \text{ is 2-relevant w.r.t. } s$$

In the previous example, $t$ may be the fact that Ryan lost faith. More generally, any move that diminishes $U(s)$ is 2-relevant w.r.t. $s$. This definition covers not only the phenomenon of trivialization [7] ("The same happened to me…"), but also any attempt to diminish $C_w(s)$ by simplifying the generation scenario (*i.e.* by providing an explanation).

## 5 Examples

### 5.1 The 'Nude Model' Story

The story in figure 3 is adapted from a spontaneous conversation analyzed by Neal Norrick [17]. The story is about a fortuitous encounter with a model who was previously seen posing in the nude. Elements indicated in bold face are commented on below.

> B: It was just about **two weeks ago**. And then we did some figure drawing. Everyone was kind of like, "oh my God, we can't believe it." We- y'know, **Midwest College**, y'know,
> […]
> B: like a … **nude models** and stuff. And it was really weird, because then, like, **just last week**, we went downtown one night to see a movie, and we were sitting in [a restaurant], like downtown, waiting for our movie, and **we saw her** in the [restaurant], and it was like, "that's our model" (laughing) **in clothes**
> A: (laughs) Oh my God.
> B: we were like "oh wow." It was really weird. **But it was her**. (laughs)
> A: Oh no. Weird.
> B: I mean, that's weird when you run into somebody **in Chicago**.
> A: yeah.

**Fig. 3.** The Nude Model story (after [17]).

The mention "just last week" is not here by chance. Recent events are simple to describe, what makes them more likely to appear unexpected. Intuitively, the story is better so, than if the time reference had been "one year ago". Formula (1) explains why. If $a$ is the typical duration of this kind of episode, then the complexity of locating the event at time location $T$ in the past amounts to $\log_2(T/a)$ bits. Formula (1) predicts logarithmic recency effects: unexpectedness varies as $-\log_2(T)$. If B had not made temporal location explicit, she would have implicitly meant "at some point in my life". The mention "just last week" is thus relevant according to (5).

When B locates the initial episode by mentioning "two weeks ago", she also makes a relevant move. This story is about a coincidence. It depicts two situations in which B has encountered the model. When two independently generated situations $s_1$ and $s_2$ bear some resemblance, the joint event is unexpected. It has been observed that Kolmogorov complexity is the right tool to quantify the intensity of coincidences ([11] p. 967). Coincidences can indeed be shown to be unexpected, according to definition (1). Let us first observe that generation complexity captures the idea that the coinciding situations are independent:

$$s_1 \text{ and } s_2 \text{ are independent if } \quad C_w(s_1 \wedge s_2) = C_w(s_1) + C_w(s_2) \tag{6}$$

We get:

$$U(s_1 \wedge s_2) \geq C_w(s_1) + C_w(s_2) - C(s_1) - C(s_2 \mid s_1) \tag{7}$$

We see from (7) that the resemblance between $s_1$ and $s_2$ is crucial for producing unexpectedness, as it makes $C(s_2 | s_1)$ smaller than $C(s_2)$. In particular, the temporal location of $s_1$ may be used to locate $s_2$. If $\Delta$ is the temporal distance between $s_1$ and $s_2$, then $s_2$ can be located from $s_1$ using only $\log_2(\Delta/a)$ bits. The economy in the description generates unexpectedness. The mention "two weeks ago" is thus essential.

The model's nudity is essential to the story. With a dressed model, the story would be much poorer indeed. This simple property, having been seen naked, makes the model simple to B's eyes, in two ways. The model belongs to the restricted set of the $n$ people who were naked in B's company. Her complexity is at most $C(model|naked) \leq \log_2(n)$. But nudity makes the model simple in another way. She was that (unique) person who was seen posing in the nude in a Midwest College.

The mention "Midwest College" indeed contributes to the story's unexpectedness. B makes it explicit that figure drawing with a nude model is a truly exceptional situation in such an institution. Interest would lessen if B had been attending an art school with regular life drawing. This time, unexpectedness is due to the difficulty of generating the situation. Nude models do not belong to Midwest colleges. Generating a situation that contradicts this statement is as complex as the statement's mutability is low (see section 5.2).

The obvious mention "in clothes" contributes to the complexity contrast: the minimal scenario that allowed B to see the same person in public twice, once naked, once in clothes, cannot be simple.

The actual presence of the model in the restaurant ("But it was her") is crucial. The description complexity of the dressed person would have been significantly larger if she just looked like the nude one. The mention "we saw her" is relevant in a similar way. B reports the event as a first-hand story. The same anecdote would appear less interesting if it had happened to one of B's friend C. The complexity of C would have been subtracted from $U(s)$.

B feels the necessity of mentioning a fact which is also obvious to her interlocutor, when she specifies "in Chicago". The size of the city, measured for instance by the number $N$ of its buildings, matters here. If the second encounter is supposed to be generated through a lottery, then generating the presence of the model in a specified place amount to $\log_2(N)$. So the relevance increases as $\log_2(N)$ (note that if there were $k$ people in that place, then a term $-\log_2(k)$ comes from the description side, due to the indeterminacy).

We observe that by equating relevance with simplicity (complexity drop) and with unexpectedness, we are able to account for the various parameters that control interest in this story. This is a non-trivial and falsifiable result, which is in line with the importance of algorithmic complexity in cognitive computations.

### 5.2 The 'rally' discussion

Let's consider now relevance within a discussion. The discussion in figure 4 occurred between French students. F will graduate in a few months and will no longer be a student next year. When F claims he wants to participate in the rally next year, G points to an inconsistency. The reminder of the discussion is about whether the contradiction is real or not.

> F- This rally, wonderful! I'm ready to come back from Toulouse next year to participate.
> G- Yes, but it is only for students, isn't it?
> T- No, no, it's open to everyone.
> F- There were people from Arcade!
> G- Yes, but they were sponsors!

**Fig. 4.** The Rally discussion (translated from French).

In the discussion of figure 4, G draws attention to a logical clash between three propositions: $\neg f_1$ = 'not being a student', $f_2$ = 'participate in the rally' and $f_3$ = 'the rally is only for students' ($\neg$ refers to negation). As we will see, the effect of G's first utterance is to increase the generation complexity of $f_2$, and so to make the situation unexpected. This is what makes G's move relevant.

Several links exist between logic and generation complexity. Some are listed below ($\supset$ refers to implication):

$$C_w(a \vee b) = \min(C_w(a), C_w(b)) \qquad (8)$$

$$C_w(a \wedge b) \leq C_w(a) + C_w(b) \qquad (9)$$

$$\text{If } (a \supset b), \text{ then } C_w(a) \geq C_w(b) \qquad (10)$$

In our example, the incompatibility between $\neg f_1$, $f_2$ and $f_3$ can be rewritten: $f_2 \supset (f_1 \vee \neg f_3)$. We get:

$$C_w(f_2) \geq C_w(f_1 \vee \neg f_3)$$

and thus:

$$C_w(f_2) \geq \min(C_w(f_1), C_w(\neg f_3)) \qquad (11)$$

G's point is that both $f_1$ (F will still be a student next year) and $\neg f_3$ (the rally is open to anyone) are hard to generate, and so is $f_2$ (F's participation). Due to the large value of $C_w(f_2)$, $f_2$ appears unexpected and G's point is relevant.

Generation complexity can be linked to the notion of *mutability* [18]. Facts about the world are memorized with 'necessity' values that are due to beliefs. I believe that my bank account balance is positive and I believe that the capital city of France is Paris, but the former belief is more mutable than the latter. To measure mutability, we have to consider the least complex combination of circumstances that can change the observer's belief toward a proposition *f*. For my bank account to be in the red right now, I must imagine an abnormal expense that I would have forgotten, or some computer error, or that my salary has been seized by some unknown court decision. Let's call *H(f)* the least complex scenario that can produce *f*. We may write:

$$C_w(f) = C_w(H(f)) \tag{12}$$

The mutability $M(f)$ of $f$ can be defined as:

$$M(f) = -C_w(H(\neg f)) = -C_w(\neg f) \tag{13}$$

(note that mutability is always negative). When $H(\neg f)$ is complex, $\neg f$ is complex to generate and the fact $f$ is not mutable ($M(f) \ll -1$). $M(f)$ might be retrieved from memory, or directly computed by finding out the most convincing (or least unconvincing) scenario $H(\neg f)$. It can be also inherited through (10) which can be rewritten $C_w(\neg b) \geq C_w(\neg a)$:

$$\text{If } (a \supset b), \text{ then } M(a) \geq M(b) \tag{14}$$

The conversation of figure 4 offers an example of 2-relevance. F's second utterance: "There were people from Arcade", is meant to refute $f_3$. Those people work in a company and are not students, and yet they were among participants. So $(\neg f_1 \wedge f_2)$ is easy to generate. Since $f_3 \supset \neg (\neg f_1 \wedge f_2)$, we get from (14) that $M(f_3) \geq -C_w((\neg f_1 \wedge f_2))$. $f_3$ is therefore highly mutable. Relation (11) no longer constrains $C_w(f_2)$ to be large. F's second utterance is thus 2-relevant.

More generally, any attempt to solve a problematic fact $f$ will be 2-relevant. The solution may be a belief revision or a new and simpler scenario $H(f)$. In any case, it leads to a diminution of $C_w(f)$, which may be named 'compression', as in Gregory Chaitin's aphorism "comprehension is compression" [19] (see [11], section 7.3 for a review of ideas about compression and explanation).

## 6  Discussion

What precedes is an attempt to account for the phenomenon of relevance in terms of complexity. The principal departure from standard algorithmic theory is that we distinguish between generation complexity and description complexity, and that all computations are performed on specific 'machines'. This approach offers numerous advantages. For instance, it predicts that the relevance of an event occurring at distance $d$ from the observer varies like $2 \times \log_2(d)$ [12]. It also predicts that for an object $s$ randomly taken from a class $r$ to be relevant, both the class and the reason $f$ that makes $s$ unique must be simple:

$$C(s) \leq C(r) + C(f \mid r) + C(s \mid r \wedge f)$$

If there are $N$ objects in the class, then $C_w(s) = \log(N)$. If we assume uniqueness of $s$ knowing $r$ and $f$, $C(s \mid r \wedge f) = 0$ and:

$$U(s) \geq \log_2(N) - C(r) - C(f \mid r) \tag{15}$$

Relation (15) may be tested by its predictions of relevance in a collection of records such as the Guinness Book. It also open the way to automated news selection.

Another advantage of the algorithmic approach to relevance is that it is closer to the possibility of implementation than alternative definitions, such as [14]. As illustrated in table 1, description complexity values can be assessed through various practical means. Generation complexity values can be computed using (6), (8), (10) to combine the parameter settings of simple machines such as lotteries.

One limitation of the above definition of relevance is that it does not take emotion into account. The emotional scale $E(s)$ on which an event or discussion topic $s$ is placed is an essential ingredient of relevance. It is not equivalent to speak about the loss of people's life or the loss of ten Euros. Relevance $I(s)$ is a function of the emotional scale and of unexpectedness:

$$I(s) = F(E(s), U(s)) \qquad (16)$$

Relation (16) means that once the emotional scale is determined, emotional intensity (and thus relevance) is entirely controlled by unexpectedness. $F$ is an increasing function of its two arguments. Determining the nature of $F$ remains a problem and is a topic of future investigations.

We are aware of the fact that the notions developed in this paper may benefit from a formal description of the generation machine and of the observation machine. This research program is motivated by the assumption that the human brain is sensitive to algorithmic complexity [2], even at higher cognitive levels where relevance is processed. The results already obtained are encouraging. They show that algorithmic complexity is not bound to deal with theoretical computer science and prove mathematical theorems, but can also be used to model particular machines such as the human mind. We expect that other important aspects of cognitive processes will be analyzed using an algorithmic complexity approach, and that these new insights will lead to implementations.